\pdfoutput=1
\documentclass[11pt]{article}
\usepackage{xcolor}

\usepackage{acl}

\usepackage{times}
\usepackage{latexsym}

\usepackage[T1]{fontenc}
\usepackage[utf8]{inputenc}

\usepackage{microtype}

\usepackage{inconsolata}

\usepackage{amsmath,amssymb,amsfonts}
\usepackage{mathtools, cuted}
\usepackage{graphicx}
\usepackage{textcomp}
\usepackage{cleveref}
\usepackage{glossaries}
\usepackage{hyperref}
\usepackage{booktabs}
\usepackage{algorithm}% http://ctan.org/pkg/algorithms
\usepackage{algpseudocode}% http://ctan.org/pkg/algorithmicx
\usepackage{xspace,mfirstuc,tabulary}
\usepackage{multirow}
\usepackage{balance}
\usepackage{enumitem}
\usepackage{listings}
\usepackage{subfig}
\usepackage[subfigure]{tocloft}
\usepackage{dblfloatfix}
\usepackage{makecell}
\usepackage{float}
\usepackage{multicol}
\usepackage{seqsplit}

\makeatletter
\def\maketag@@@#1{\hbox{\m@th\normalfont\normalsize#1}}
\makeatother

\makeatletter
\newcommand*{\glsplainhyperlink}[2]{%
  \colorlet{currenttext}{.}% store current text color
  \colorlet{currentlink}{\@linkcolor}% store current link color
  \hypersetup{linkcolor=currenttext}% set link color
  \hyperlink{#1}{#2}%
  \hypersetup{linkcolor=currentlink}% reset to default
}
\let\@glslink\glsplainhyperlink
\makeatother
\glsdisablehyper

\DeclareMathAlphabet\mathbfcal{OMS}{cmsy}{b}{n}

\author{
  Pascal Wullschleger$^{*,\dagger}$,
  Majid Zarharan$^\diamond$,
  Donnacha Daly$^\dagger$\\
  \bf{Marc Pouly$^\dagger$,
  Jennifer Foster$^*$} \\\\
  $^*$ Hamilton Institute, Maynooth University, Ireland\\
  $^\diamond$ School of Computing, Dublin City University, Ireland\\
  $^\dagger$ Lucerne School of Computer Science and IT, Switzerland\\\\
  \texttt{pascal.wullschleger@hslu.ch}
}

\definecolor{codegreen}{rgb}{0,0.6,0}
\definecolor{codegray}{rgb}{0.5,0.5,0.5}
\definecolor{codepurple}{rgb}{0.58,0,0.82}
\definecolor{backcolour}{rgb}{0.97,0.97,0.97}
\lstdefinestyle{verbatim}{
  backgroundcolor=\color{backcolour}, commentstyle=\color{codegreen},
  keywordstyle=\color{magenta},
  numberstyle=\tiny\color{codegray},
  stringstyle=\color{codepurple},
  emphstyle={\color{cyan}},
  basicstyle=\fontsize{8}{8}\selectfont\ttfamily,
  breakatwhitespace=false,         
  breaklines=true,                 
  captionpos=b,                    
  keepspaces=true,                 
  numbersep=5pt,                  
  showspaces=false,                
  showstringspaces=false,
  showtabs=false,                  
  tabsize=2,
  breakindent=4ex
}

\lstdefinelanguage{prompt}
{
  alsoletter=-:,
  morekeywords={
    hyponym,
    hypernym,
    hyponym_article,
    hypernym_article,
    hypernym_plural 
  },
  emph={---},
  sensitive=true, % keywords are not case-sensitive
  morecomment=[l]{//}, % l is for line comment
  morestring=*[d]{```} % defines that strings are enclosed in triple backticks
}

\title{Reference-Free Evaluation of Taxonomies}
\date{August 2023}

\newacronym{LLM}{LLM}{Large Language Model}
\newacronym{NLP}{NLP}{natural language processing}
\newacronym{RAG}{RAG}{retrieval augmented generation}
\newacronym{DSP}{DSP}{Demonstrate-search-predict}
\newacronym{CoT}{CoT}{Chain-of-thought}
\newacronym{NLI}{NLI}{Natural Language Inference}
\newacronym{WPS}{WPS}{Wu \& Palmer Similarity}
\newacronym{CSC}{CSC}{Concept Similarity Correlation}
\newacronym{GRU}{GRU}{Gated Recurrent Unit}
\newacronym{SP}{SP}{Semantic Proximity}
\newacronym{WOS}{WOS}{Web Of Science}
\newacronym{MN-DS}{MN-DS}{Multilabeled News Dataset}
\newacronym{DBP}{DBP}{DBPedia}

\begin{document}

{\makeatletter\acl@finalcopytrue
  \maketitle
}

\begin{abstract}
We introduce two reference-free metrics for quality evaluation of taxonomies in the absence of labels.\ 
The first metric evaluates robustness by calculating the correlation between semantic and taxonomic similarity, addressing error types not considered by existing metrics.\ 
The second uses Natural Language Inference to assess logical adequacy.\ 
Both metrics are tested on five taxonomies and are shown to correlate well with F1 against ground truth taxonomies. 
We further demonstrate that our metrics can predict downstream performance in hierarchical classification when used with label hierarchies.
\end{abstract}

\section{Introduction}
Taxonomies are used to classify items, ideas or 
organisms based on shared characteristics. They help organize information, making it easier to find, understand and manage. 
% In the food industry, for example, taxonomies play a critical role in the development of new 
% recipes and their adaptation to changing culinary trends, dietary requirements, and sustainability objectives. 
Hierarchical classification, for example, enables systems to progressively categorize items from broad groups to specific subcategories, thereby simplifying complex classification tasks by leveraging the taxonomy of labels \cite{electronics13071199}. Likewise, taxonomies can be used to enrich features for machine learning 
%models
\cite{10.1145/3097983.3098126, 10.1145/3269206.3271701, 10020339} or improve zero-shot classification \cite{10.1145/3442381.3450042, ijcai2019p845, KR2020-87}.

Taxonomies have traditionally been created manually, 
%by humans
 and,  
 more 
 recently,  partially or fully automatically. Such automated taxonomy extension or generation methods address the challenges posed by ever-increasing volumes of unstructured 
 data.
 %information. 
 By leveraging machine learning, systems can identify emergent categories, detect previously unrecognized relationships, and update or create taxonomic structures in real time with minimal human effort \cite{velardi-etal-2013-ontolearn, jurgens-pilehvar-2016-semeval, bordea-etal-2016-semeval, Zhang_Song_Zeng_Chen_Shen_Mao_Li_2021}.

Taxonomy completion methods are  usually evaluated by comparing the resulting taxonomies against ground truth taxonomies 
\cite{10.1145/3366423.3380271, 10.1145/3485447.3511943,liu-etal-2021-temp,Zhang_Song_Zeng_Chen_Shen_Mao_Li_2021,xu-etal-2023-tacoprompt,10.1145/3366423.3380132, 10.1145/3589334.3645584, zeng-etal-2025-codetaxo}.
 In practice, however, there is no ground truth taxonomy to compare against when generating or completing taxonomies, because, if the ground truth taxonomy was available, there would be no need to expand or generate it. 
 A potential solution is to use reference-free evaluation. We propose two reference-free measures, each targeting a particular aspect of taxonomic quality. 

  % A second problem with comparing against a ground truth taxonomy, is that, even when such a taxonomy is available, there can be multiple valid taxonomies for a given set of leaf concepts and one reference taxonomy is unlikely to cover all valid relationships. Consider the examples in Fig.???. Both taxonomies are valid, although one might be better than the other depending on the application.
% To our knowledge, efforts 
% in reference-free taxonomy evaluation
% have been limited to work by \citet{langlais-gao-2023-rate}. 

% The point is, that taxonomies are not usually exhaustive and there are various approaches to classify the same concepts hierarchically.   

\begin{figure}[!t]
\centering
\subfloat[Robustness]{\label{fig:sp_toy_rob}
\includegraphics[width=0.15\textheight]{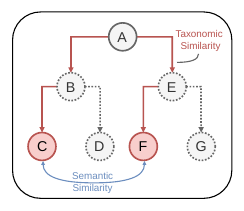}}\hfil
\subfloat[Logical Adequacy]{\label{fig:sp_toy_la}
\includegraphics[width=0.15\textheight]{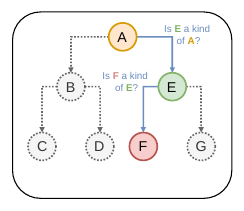}}
\caption{Core ideas of our reference-free measures. We correlate semantic and taxonomy similarity to measure robustness and check parent-child edges to assess logical adequacy of the taxonomy.}
\end{figure}

We build on established taxonomy quality criteria, particularly robustness \cite{https://doi.org/10.1111/exsy.13098}, but find existing automated measures do not account for non-leaf mutations, overlook error severity or heavily rely on priors within the backbone model's training data. In addition, criteria aimed at human-curated taxonomies overlook logical adequacy, such as whether a child concept is truly a subtype of its parent, presumably because 
%humans 
people
rarely make such errors. Consequently, we suggest two 
%basic 
properties for evaluating taxonomy quality in a reference-free setting: the correlation of semantic and taxonomic similarity (\textit{robustness} -- see Fig.~\ref{fig:sp_toy_rob}) and the recognition of \texttt{is-a} parent-child edges (\textit{logical adequacy} -- see Fig.~\ref{fig:sp_toy_la}). 
%
%To sum up, the contributions of this study are novel metrics for evaluating the robustness and logical adequacy of taxonomies in a reference-free fashion. 

%In a comparison of five different taxonomies, 
We %investigate the potential of 
evaluate
our 
%proposed 
metrics on five 
%different 
taxonomies across 
%the food, lexical, and medical 
various
domains. By 
%intentionally 
degrading ground truth taxonomies, we demonstrate that our metrics align with the level of deterioration, outperforming existing metrics.
%In a similar fashion, 
Similarly, 
we demonstrate that 
%our metrics 
they
outperform
%show superior correlation to 
existing metrics in an extrinsic evaluation of taxonomy quality using the downstream application of hierarchical classification. Our source code is available on GitHub\footnote{\href{https://github.com/wullli/reference-free-taxonomy-eval}{https://github.com/wullli/reference-free-taxonomy-eval}}. 
%Moreover, we show that applying our metrics to the label taxonomy used in a hierarchical classification task can predict hierarchical 
%classification results.

\newcommand{\T}{\mathcal{T}}
\newcommand{\E}{\mathcal{E}}
\newcommand{\V}{\mathcal{V}}
\newcommand{\C}{\mathcal{C}}
\newcommand{\mL}{\mathcal{L}}

\section{Background}
\subsection{Preliminaries}
%\subsection{Taxonomy}
Following \citet{10.1145/3447548.3467308}, a taxonomy $\T = (\E, \V)$ is a 
directed acyclic graph 
%(DAG)
with edges $\langle c_i, c_{i+1}\rangle \in \E$ pointing 
from a parent concept $c_i \in \V$ to a child concept $c_{i+1} \in \V$.\
Edges represent hypernym-hyponym pairs, where the child concept is the 
least detailed but different specialization of the parent concept.\ A placement of a concept is called a triplet $(c_i, c_{i+1}, c_{i+2})$, where $c_{i+1}$ 
is the query concept that is placed as a child of $c_i$ and  parent of $c_{i+2}$.\
Following \citet{10.1145/3366423.3380271}, we add a pseudo-leaf and a pseudo-root to $\V$ 
to accommodate concepts without parents or children. If $c_{i+1}$ is a leaf, $c_{i+2}$ is the pseudo-leaf and if instead $c_{i+1}$ is the root, then $c_{i}$ is the pseudo-root. To compare a 
%(possibly generated) 
taxonomy to a ground truth, precision/recall/F1 are calculated over triplets.
%For example: a true positive would mean, that the triplet exists in both the custom and the gold standard taxonomy 

%\subsection{Taxonomic Similarity}
As a measure of taxonomic similarity, we use the well-established \gls{WPS} \cite{Wu1994-yf}.\
%, which is commonly known for its application as a similarity score with WordNet \cite{Fellbaum2019-dh}. 
%Let $p(c_t)$ denote the path from the root concept $c_r$ to a target concept $c_t$ as a tuple of nodes $(c_r, ..., c_t)$. 
Let $p(c_k) = \langle c_0, ..., c_k\rangle$ denote the path from the root concept $c_0$ to a target concept $c_k$.\
Furthermore, let $\mathtt{lca}(p(c_a), p(c_b))$ denote the depth of the least common ancestor of the paths $p(c_a)$ and $p(c_b)$. Then the \gls{WPS} (Eq. \ref{eq:wps}) 
%represents 
is
the similarity between concepts $c_a$ and $c_b$ in the range $(0, 1]$, with $1$ meaning that they are the same node.

\begin{small}
\begin{eqnarray}
W_{c_a c_b} &=& \frac{2 \cdot \mathtt{lca}(p(c_a), p(c_b))}{\lvert p(c_a)\rvert + \lvert p(c_b)\rvert}
\label{eq:wps}
\end{eqnarray}
\end{small}

\subsection{Taxonomy Quality Attributes} 
% The most commonly reported quality attributes used to assess taxonomies are usefulness and applicability. 
% Evaluation typically involves applying the taxonomy to a sample dataset to determine how effectively it classifies the data \cite{Gimpel2018-ms,Gibbs2016-zb,Williams2008-my}. However, 
%In their compendium of taxonomy quality attributes, 
% \citet{Unterkalmsteiner2023-wt} regard these factors as placeholders for a more essential set of attributes. %The authors
\citet{https://doi.org/10.1111/exsy.13098} consolidate various names for taxonomy quality attributes found in prior work and identify a measurable minimal, well-defined set --- comprehensiveness, robustness, conciseness, extensibility, explanatory, mutual exclusiveness, reliability (criteria definitions in App.~\ref{app:sec:criteria}). 
%They evaluate this set on six taxonomies across three domains .
%to demonstrate the utility.
%of their proposed measurements.
%
%
%In this study, 
We focus on \textit{robustness}, which represents how well a taxonomy can tell things apart, meaning how clearly the concepts in a taxonomy represent different ideas 
%(orthogonality) 
and how closely related sibling concepts are. 
%(cohesiveness)
Robustness is independent of the application of the taxonomy and depends only on the concept space, since it describes an intrinsic quality. This makes it suitable for automated evaluation in a reference-free setting. 

We also observe that prior work does not account for the need to verify \texttt{is-a} relationships.
For instance, a parent \emph{beverage} and a child \emph{bread} would constitute an invalid relationship,
since bread is not actually a type of beverage.
%of the  property.
%in taxonomies. 
%We assume this stems from the fact that the \texttt{is-a} edge is a fundamental characteristic of taxonomies, making 
While we assume such erroneous edges to be rare in human-created taxonomies, this does not necessarily hold for automatically generated taxonomies.
%and so it needs to be assessed. 
We refer to this property as \textit{logical adequacy}.

% Perhaps the most closely related work to ours is by \citet{Langlais2023-nz}, who propose rating parent-child edges in a taxonomy by generating valid parent concepts for a given child using masked language model prompts and assessing whether the true parent appears in the set of valid parents. While this approach aligns with our work as a step toward intrinsic taxonomy evaluation, it does not assess the overall quality of the hierarchy as robustness would. Additionally, errors at higher levels of the taxonomy do not propagate downward to descendant concepts.

\section{Methodology}

This section outlines our methods for evaluating taxonomy robustness and logical adequacy. We also apply these and prior metrics to a toy example.
% for illustration purposes. 

\subsection{Robustness}
\label{sec:method_robustness}
\begin{figure}[t!]
\centering
\subfloat[Original]{\label{fig:sp_toy_orig}
\includegraphics[height=0.085\textheight] {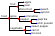}}
\subfloat[Mutation 1 (leaf \textit{apple})]{\label{fig:sp_toy_2}
\includegraphics[height=0.085\textheight] {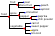}}\\
\subfloat[Mutation 2 (non-leaf \textit{stone fruit})]{\makebox[\linewidth][c]{\label{fig:sp_toy_1}
\centering\includegraphics[height=0.085\textheight] {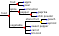}}}\\
\caption{Toy taxonomy to illustrate different mutations that a robustness metric should be sensitive to. Leaf nodes are marked in blue, mutated nodes in yellow.}
\label{fig:sp_toy_expl}
\end{figure}

% We suggest the following two (non-exhaustive) scenarios to which taxonomy quality measures should be sensitive, and propose mutations to test the scenarios on a toy example (see Fig. \ref{fig:sp_toy_expl}).
Fig. \ref{fig:sp_toy_expl} shows two basic mutations of a toy taxonomy to which quality measures should be sensitive:
\begin{enumerate}[itemsep=0em]
\item \textbf{Misclassified leaf concepts:} Move a leaf by randomly choosing a new parent (Fig. \ref{fig:sp_toy_2}).
\item \textbf{Misclassified non-leaf concepts:} Move a non-leaf and its descendants by randomly choosing a new parent. (Fig. \ref{fig:sp_toy_1}).
\end{enumerate}

\citet{https://doi.org/10.1111/exsy.13098} propose \gls{SP} as a robustness metric that measures the rate of intruders in groups of siblings.\ They check if the minimum inter-group similarity is larger than the minimum extra-group similarity and average over all groups (see App. \ref{app:sec:semprox}).\
The SP metric disregards the hierarchical nature of taxonomies. It essentially treats groups of leaves as clusters, but does not consider that they could be nested in a hierarchical structure.\ We expect SP to be insensitive to misclassifications of non-leaf nodes (Fig. \ref{fig:sp_toy_1}) since it only looks
at groups of leaf concepts to calculate a score.

To mitigate this, we propose 
%an alternative robustness measure, 
\textit{\gls{CSC}}, which is based on the assumption that the semantic distance between concepts
%names or descriptions 
and their taxonomic 
distance should correlate.\ We measure the \acrshort{WPS}, $W_{c_i c_j}$, between concepts $c_i$ and $c_j$ using the paths from the root to the concepts~(Eq.~\ref{eq:wps}). 
Next, for each concept in the taxonomy, we define its semantic representation as the 
embedding of the concept description.\ Our \gls{CSC} score is the Kendall rank correlation $\tau(.)$ of the 
semantic (cosine) similarity $S_{c_i c_j}$ between concept representations and the taxonomic
similarity $W_{c_i c_j}$ (Eq. \ref{eq:cdc}). Thus, we assume that the relationship between taxonomic and semantic similarity is monotonically 
increasing in a sensible taxonomy, i.e. if semantic similarity increases, so does taxonomic similarity. We calculate Kendall’s $\tau$ using all pairs of concepts in the taxonomy (exhaustively). 
%should also 
%increase.

\begin{small}
\begin{equation}
\label{eq:cdc}
\text{CSC} := \tau\left(S_{c_i c_j}, W_{c_i c_j}\right)
\end{equation}
\end{small}

\subsection{Logical Adequacy}
%Since a robustness measure does not necessarily respect that parent-child edges need to be of the type \texttt{is-a} in a taxonomy, we introduce a second measure to assess this. 

%\paragraph{Probabilistic Interpretation}
Since robustness does not ensure valid parent-child edges, we 
%subsequently 
propose a logical adequacy measure. %to assess adequacy. 
Let us assume an external object that we want to classify using a taxonomy. Then, a classification process is a walk on the taxonomy graph represented as a 
tuple of edges $C =  (\langle c_0, c_1 \rangle, \ldots, \langle c_{k-1}, 
c_k\rangle )$ that connect the root concept $c_0$ to any concept $c_k$ under 
which the object is classified, with \( c_{i+1} = \mathtt{child}(c_i)
\) and \( c_i, c_{i+1} \in \mathcal{V} \) for \( i = 0, \dots, k - 1\). 

We want the possible classifications (walks) to be probable (according to some model) for a high quality taxonomy, i.e. the parent-child edges of the walks need to be logically adequate. For analogy, in the context of language modelling, the objective is to generate probable or likely texts, implying that a probable text is considered a good one. Similarly, in our approach, parent-child relationships must be probable for a classification to be considered good. We do not simply score each edge and average, since we want mistakes at higher levels to impact the score more than e.g., misclassified leaves deep in the taxonomy, because they affect a larger sub-graph. Thus, inner concept scores are deliberately used multiple times in the computation.

 If $A$ denotes the event of adequacy, we define the normalized probability of the classification $C$ being adequate as the geometric mean of the edge probabilities (Eq. \ref{eq:pnorm})

\begin{small}
\begin{equation}
\tilde{P}(A|C) := \left[ \prod_{i=0}^{k-1} P(A | \langle c_i, c_{i+1} \rangle)\right]^{\frac{1}{k}}
\label{eq:pnorm}
\end{equation}
\end{small}
We 
%can now 
express the adequacy of the whole taxonomy, $\tilde{P}(A)$, as the normalised probability that a randomly sampled classification in our taxonomy is adequate, which is the union probability across classifications. This results in the mean over probabilities of adequacy given the classification (Eq. \ref{eq:union}). Refer to App. \ref{app:log_ad} for further detail.

\begin{small}
\begin{equation}
\begin{split}
\tilde{P}(A) &:= \sum\limits_{C} \tilde{P}(A | C) \times P(C)
= \frac{1}{|\V|} \sum\limits_{C} \tilde{P}(A | C)\\
\end{split}
\label{eq:union}
\end{equation}
\end{small}

\paragraph{Approximating Edge Probabilities}
We estimate $P(A | \langle c_i, c_{i+1} \rangle)$, the probability of an edge being adequate, using \gls{NLI}. We rely on a flavour of \gls{NLI} that approximates a probability distribution over the classes \textit{contradicts}, \textit{neutral} and \textit{entails}, regarding whether a 
%given 
premise entails a 
%given 
hypothesis. For example, consider the edge (\textit{appetizer}, \textit{antipasto}) in a taxonomy of food items. We define the premise as the description of the child concept, e.g.,\ \textit{``antipasto is a course of appetizers in an Italian meal''} and the hypothesis as a string characterising the is-a relationship between the child and parent nodes in an edge, e.g. \textit{``antipasto is a type of appetizer''}. 
The expression $s(\langle c_i, c_{i+1} 
\rangle)$ denotes the string of the premise and hypothesis resulting from the application of the template in Eq. \ref{eq:nli_template} to the edge, where $D(c_i)$
denotes the description of the concept $c_i$, $N(c_i)$ the lemma of the concept name,
and $A(c_i)$ the appropriate article (e.g., \emph{a}, \emph{an}, \emph{the}). We abbreviate metrics based on this NLI-approximation as NLIV (NLI Verification).

\begin{small}
\begin{align}
\label{eq:nli_template}
&s(\langle c_i, c_{i+1} \rangle) = &&\\
&''D(c_{i+1}).\, A(c_{i+1}) \, N(c_{i+1}) \; \text{ is a type of } \; N(c_i)'' &&\nonumber
\end{align}
\end{small}

There are two possibilities for estimating the probability that an edge is adequate:
%(string template in Eq.~\ref{eq:nli_template}):
\begin{enumerate}
%[itemsep=0.2em]
\item \textbf{Strong:} the premise must \textit{entail} the hypothesis (\textit{NLIV-S}) as shown in Eq. \ref{eq:nliv_strong}.

\begin{small}
\vspace*{-2em}
\begin{flalign}
\label{eq:nliv_strong}
\;\;\;\;&P(A \mid \langle c_i, c_{i+1} \rangle) \approx &&\\ &\mathit{NLI}(Y = \mathtt{entails} \mid s(\langle c_i, c_{i+1} \rangle))&&\nonumber
\end{flalign}
\end{small}

\item \textbf{Weak:} the premise must \textit{not contradict} the hypothesis (\textit{NLIV-W}) as shown in Eq. \ref{eq:nliv_weak}.

\begin{small}
\vspace*{-2em}
\begin{flalign}
\label{eq:nliv_weak}
\;\;\;\;&P(A \mid \langle c_i, c_{i+1} \rangle) \approx &&\\ &1 - \mathit{NLI}(Y = \mathtt{contradicts} \mid s(\langle c_i, c_{i+1} \rangle))&&\nonumber
\end{flalign}
\end{small}
\end{enumerate}
% the description of the child concept (i.e. the premise)
% \begin{enumerate}
% \item must \textit{entail} the edge (i.e. the hypothesis). We call this \textbf{strong} NLI (\textit{NLIV-S}).
% \item must \textit{not contradict} the edge. We call this \textbf{weak} NLI  (\textit{NLIV-W}).
% \end{enumerate}
% The two
% versions are shown in Eqs. \ref{eq:nliv_strong} and \ref{eq:nliv_weak} for
% parent $c_i$ and child $c_{i+1}$. 

\paragraph{Hearst Patterns}
Instead of only relying on a single prompt template for NLI-approximation, we use the mean estimate over instances of Hearst patterns \cite{hearst-1992-automatic}, %These patterns 
which
were originally designed to extract hypernym-hyponym edges from natural language texts using regular expressions. 
%The used instances of the 
% initial two 
The first pattern is presented below (complete list in App.~\ref{app:hearst}):

\begin{small}
%\begin{itemize}[leftmargin=*]
%\itemsep0em 
%\vspace*{-0.3em}\item \textbf{Pattern (1):} %\label{eq:pattern1}
\vspace*{-0.3em}\begin{align*}
&s(\langle c_i, c_{i+1} \rangle) \in S, \; S = \{\\
&\quad ''D(c_{i+1}).\, A(c_{i+1}) \, N(c_{i+1}) \text{ is a type of } N(c_i)'', \\
&\quad ''D(c_{i+1}).\, A(c_{i+1}) \, N(c_{i+1}) \text{ is an example of } N(c_i)'', \\
&\quad ''D(c_{i+1}).\, A(c_{i+1}) \, N(c_{i+1}) \text{ is } A(c_i) \, N(c_i)'', \\
&\quad ''D(c_{i+1}).\, A(c_{i+1}) \, N(c_{i+1}) \text{ is a kind of } N(c_i)''\}
\end{align*}

% \vspace*{-0.3em}\item \textbf{Pattern (2):} \label{eq:pattern2}
% \vspace*{-0.3em}\begin{align*}
% &s(\langle c_i, c_{i+1} \rangle) \in S, \; S = \{\\
% &\quad ''D(c_{i+1}).\, A(c_i) \, N(c_i) \text{ such as } A(c_{i+1}) \, N(c_{i+1})'', \\
% &\quad ''D(c_{i+1}).\, \text{Such } N(c_i)_{\text{plural}} \text{ as } N(c_{i+1})''\}
% \end{align*}

%\end{itemize}
\end{small}

\paragraph{Parallels to Previous Work}
% Perhaps the most closely related work to ours is by \citet{Langlais2023-nz}, who propose rating parent-child edges in a taxonomy by generating valid parent concepts for a given child using masked language model prompts and assessing whether the true parent appears in the set of valid parents.
\citet{langlais-gao-2023-rate} propose rating parent-child edges %in a taxonomy 
by generating 
%valid 
parent concepts for a given child using masked language model prompts and assessing whether the true parent appears in the returned 
set. 
%of 
%valid parents. 
While they average over all parent-child pairs, 
we 
%use these approximations to 
work with complete classifications. We compare to their metric, RaTE, in our experiments.

\begin{figure}[b!]
\centering
\subfloat[Best in sample]{
\includegraphics[width=0.43\linewidth] {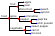}}
\subfloat[Worst in sample]{
\includegraphics[width=0.57\linewidth] {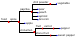}}
\caption{The best and worst toy taxonomies according to the product of \gls{CSC} and NLIV-S in a 10k sample of the toy example space.}
\label{fig:sp_toy_best}
\end{figure}%

\subsection{Toy Example}
\label{sec:toy}

The first two rows of Tab.~\ref{tab:toy_eval} show a comparison of all reference-free metrics on the toy example from Fig.~\ref{fig:sp_toy_expl}. We can see that SP is only sensitive to Mutation 1, where a leaf concept is moved, and not to Mutation 2, where a non-leaf is moved. Our proposed metrics, \gls{CSC} and NLIV, on the other hand, are sensitive to both mutations. 
% In Mutation 2, only non-leaf concepts are modified, leaving the leaves unchanged. 

For an intuition of what is considered bad by the proposed metrics, Fig.~\ref{fig:sp_toy_best} %additionally 
shows the best and worst taxonomies found by sampling 10k taxonomies in the concept space of the toy example. We do not show exhaustive search
%, even on this small space,
since even if we only consider trees (as opposed to DAGs), Cayley's formula states that the number of possible taxonomies for the toy example is already $n^{n-2} \approx 7.21 \times 10^{16}$ \cite{SHOR1995154}. The metric that we use for determining what is best and what is worst is the product of NLIV-S and CSC. The middle two rows of Tab.~\ref{tab:toy_eval} show the scores for the best and worst taxonomy according to all metrics. The taxonomy resulting in the best scores is the original toy taxonomy while the worst scores belong to a taxonomy where related concepts are far apart and hypernym-hyponym edges are invalid.

\begin{table}[t]
\resizebox{1\linewidth}{!}
{
\footnotesize\begin{tabular}{lrrrrrr}
\toprule
\textbf{Mutation} & \makecell[r]{\textbf{SP}\\$[0,1]$} &  \makecell[r]{\textbf{\acrshort{CSC}}\\$[-1,+1]$}  &\makecell[r]{\textbf{NLIV-S}\\$[0, 1]$}
&\makecell[r]{\textbf{NLIV-W}\\$[0, 1]$} & \makecell[r]{\textbf{RaTE}\\$[0, 1]$} & \makecell[r]{\textbf{CSC$\times$NLIV-S}\\$[0, 1]$}\\
\midrule
Leaf (1) & 0.4861 & 0.3745 & 0.5303 & 0.9854 & 0.8000 & 0.1986 \\
Non-Leaf (2) & 0.7805 & 0.3272 & 0.5087 & 0.9873 & 0.8000 & 0.1665 \\
\midrule
Worst in sample & 0.0625 & -0.2126 & 0.3440 & 0.9409 & 0.2667 & -0.0731 \\
Best in sample  & 0.7805 & 0.4467 & 0.5604 & 0.9909 & 0.8000 & 0.2504 \\
\midrule
Original & 0.7805 & 0.4467 & 0.5604 & 0.9909 & 0.8000 & 0.2504 \\
\bottomrule
\end{tabular}}
\caption{Results on the toy example. The value range of metric scores is indicated in brackets. 
%NLIV-S and NLIV-W denote the strong/weak versions respectively.
}
\label{tab:toy_eval}
\end{table}

\section{Metric Evaluation}

\subsection{Intrinsic Evaluation}\label{sec:intrinsic}

% Correlations
\begin{figure*}[!hb]
\centering
\includegraphics[width=0.95\linewidth]{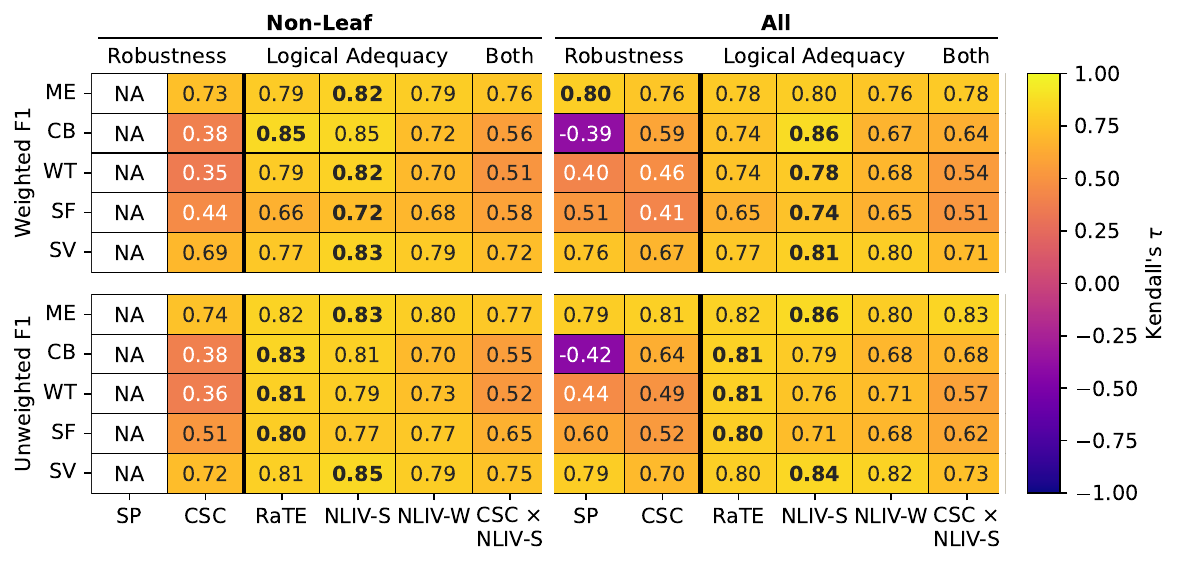}\\
\caption{Correlations between F1 and reference-free metrics, including the correlations when only mutating non-leaves. All correlations are statistically significant with $\alpha = 0.001$. We calculated F1 scores with and without class weights, assigning weights based on the number of descendants of a concept.}
\label{fig:corrs}
\end{figure*}

%We draw inspiration from the taxonomy completion literature, following 
\paragraph{Data} Following \citet{xu-etal-2023-tacoprompt} and \citet{10.1145/3485447.3511943}, we evaluate on SemEval-Food (SF), SemEval-Verb (SV), and MeSH (ME) datasets.\ SemEval-Food, the largest taxonomy from SemEval-2016 Task 13, was used to evaluate taxonomy extraction methods 
%for a given corpus 
\cite{bordea-etal-2016-semeval}. SemEval-Verb, based on WordNet 3.0 \cite{Fellbaum2019-dh}, was featured in SemEval-2016 Task 14 for evaluating taxonomy enrichment approaches \cite{jurgens-pilehvar-2016-semeval}.\ MeSH is a hierarchical vocabulary of medical terms \cite{Lipscomb2000-br}. 

In the food industry, taxonomies play a critical role in the development of new 
recipes and their adaptation to changing culinary trends, dietary requirements, and sustainability objectives. We therefore  experiment with two further taxonomies in this domain.
We extract a taxonomy from Wikidata\footnote{\url{https://www.wikidata.org/}} (WT) by using \texttt{Food (Q2095)} as the root and retrieving all descendants (see App.~\ref{app:sec:wikitax}), and we use a proprietary taxonomy from a major food market chain employed in recipe development, which we call the \textit{CookBook} (CB) taxonomy.\footnote{Both taxonomies are available 
with our source code at \href{https://github.com/wullli/reference-free-taxonomy-eval}{https://github.com/wullli/reference-free-taxonomy-eval}.} 
%upon publication.} 
For details on the datasets, see Tab.~\ref{tab:tax_stats}.

\setlength\arrayrulewidth{1pt}
\begin{table}[t!]
    \centering
    \resizebox{\linewidth}{!}
    {

    \footnotesize\begin{tabular}{lrrrrrr}
        \toprule
         \textbf{Dataset} & $|\mathbfcal{V}|$ & $|\mathbfcal{E}|$& \textbf{$\mathbf{D}$} & \textbf{$\mathbf{|L|}$} & \textbf{$\mathbf{\frac{|L|}{|\mathbfcal{V}|}}$} & $\mathbf{B}$\\ 
         \midrule
        MeSH (ME) & 9,710 & 10,496 & 11 & 5,502 & 0.57 & 3.88 \\
        CookBook (CB) & 1,985 & 1,984 & 4 & 1,795 & 0.90 & 10.44 \\
        Wikitax (WT) & 941 & 973 & 7& 754 & 0.80 & 5.20 \\
        SemEval-Food (SF) & 1,486 & 1,576 & 9 & 1,184 & 0.80 & 5.08 \\
        SemEval-Verb (SV) & 13,936 & 13,407 & 13 & 10,360 & 0.74 & 4.12 \\
        \bottomrule
    \end{tabular}
    }
  \caption{Benchmark taxonomy statistics. $|\mathcal{V}|$, $|\mathcal{E}|$, $D$, $|L|$, $\frac{|L|}{|\mathcal{V}|}$, $B$ represent the concept number, edge number, depth, leaf number, leaf ratio  and  branching factor %of the taxonomy.
  }\vspace*{-1.5em}
  \label{tab:tax_stats} 
\end{table}

\paragraph{Pre-Trained Models}
As an \gls{NLI}-model, we use
BART-L \cite{lewis-etal-2020-bart}, which used the same pre-training data as \citet{DBLP:journals/corr/abs-1908-08345}, including BooksCorpus 
%(800M words) 
\cite{10.1109/ICCV.2015.11}, English Wikipedia 
%(2,500M words, 
%only text, 
%(no meta-data and tables) 
(text only)
and various news-article datasets, and was then post-trained for zero-shot \gls{NLI} on the MultiNLI dataset \cite{williams-etal-2018-broad}. NLI models are only used in zero-shot mode in our metrics, thus we validate them only in this mode. Zero-shot is also the most challenging setting and will therefore provide a lower bound for performance. 
For semantic similarity, we use MiniLM-L6,
%in all experiments unless stated otherwise, 
which is a sentence-transformer  that generates embeddings of the mean over non-padding token representations \cite{reimers-2019-sentence-bert}. See App.~\ref{app:model_ids} for the full model identifiers.
%for both NLI and sentence-transformer models.}
\paragraph{Metric Evaluation} To evaluate the proposed metrics, we conduct an empirical validation using our benchmark taxonomies, which are assumed to be of high quality and to reflect human judgment. We degrade %versions of 
all taxonomies by randomly relocating sub-graphs of the original taxonomies according to the mutations described in Sec.~\ref{sec:method_robustness}.\ To this end, we mutate between 2\% and 32\% of the taxonomy nodes and calculate F1 (against the original taxonomy), NLIV and CSC (our proposed metrics) as well as RaTE and SP (previous metrics) every $2^x$\% mutations with $x = 1, \dots, 5$. Depending on the taxonomy, we sample between 50-100 different degradations, resulting in 250-500 degraded taxonomies.\ 
We then calculate the correlation between F1 and CSC/NLIV-S/NLIV-W/RaTE/SP. Note that mutations are sequential and a subsequent mutation can affect a previously mutated subgraph.

We present two mutation settings. In the \textit{non-leaf} setting, we only move sub-graphs of inner concepts (not leaves). In the \textit{all} setting, we allow any concept's sub-graph to be moved. Our NLIV metrics assume that mistakes higher in the taxonomy are more severe than mistakes at lower levels, since they affect more descendants. Thus, an unweighted F1-score is biased against our metric. For completeness, we calculate F1 scores weighted by the number of descendants of a node in addition to an unweighted score.

\paragraph{Results}
Fig.~\ref{fig:corrs} shows Kendall's $\tau$ between the F1 score against the ground truth taxonomy and the reference-free metric scores.\ Correlations are undefined (NA) when the metric predicts the same score for all mutated taxonomies. For instance, when mutating only non-leaf concepts, \gls{SP} remains constant and the correlations are thus undefined (NA) for all datasets. 
Our metrics, CSC and NLIV-S, are strongly associated with F1 across datasets. For leaf nodes, SP performs effectively except on CookBook, likely due to its higher branching factor (see Tab. \ref{tab:tax_stats}). SP demands that intra-group similarity be below extra-group similarity per leaf, but larger leaf groups increase intruder risk due to lower cohesion. Although CSC$\times$NLIV-S has a weaker correlation than NLIV-S alone, it helps avoid flat taxonomies. Our degraded taxonomies are unlikely to be flat, given an expected branching factor of 2 for non-leaves in the limit \cite{06f20037-d357-33ec-ba7e-06e6606e6bd0}. NLIV-W has a weaker correlation with F1, likely because the model does not always perceive invalid edges as contradictory. RaTE shows consistently high correlations, but, as expected, falls short of NLIV-S in the weighted F1. RaTE also calculates scores using only a subset of the taxonomy, since for most concepts the true parent is not actually in the top-$k$ and the score is simply zero over this subset ($\sim50\% -90\%$, 
%depending on the taxonomy, 
see Fig.~\ref{fig:scatter_all_weighted} in App.~\ref{app:sec:extended}).

\subsection{Extrinsic Evaluation}

We next evaluate how our metrics correlate with performance  on a downstream task.\
We assume that the quality of a taxonomy can be characterised by how well external objects can be classified into it, i.e. how well it performs in a hierarchical classification setting. 
For example, in hierarchical text classification, a text must be categorised into predefined hierarchically organised classes,
with general classes at the higher levels and detailed classes at lower levels (see Tab.~\ref{tab:dbp_example}).
This task is well-suited for extrinsic evaluation because a correct class hierarchy (taxonomy) is crucial in preventing error propagation.
That is, since high-level predictions influence low-level ones, a low-level target misplaced in the hierarchy is harder to predict.

\paragraph{Data}
We experiment with three  hierarchical classification %benchmark 
datasets, each accompanied
by a label taxonomy. 
For data statistics, refer to
Tab.~\ref{tab:downstream_stats}. \gls{WOS} includes abstracts of scientific articles that
must be classified into their respective topic domains (level 1) and areas (level 2) \cite{kowsari2017HDLTex}. The \gls{MN-DS} is a collection of news articles gathered over more than a year from an academic news search engine \cite{data8050074}. The \gls{DBP} dataset\footnote{\url{https://huggingface.co/datasets/DeveloperOats/DBPedia_Classes}} is based on the DBPedia project \cite{10.1007/978-3-540-76298-0_52}, which aims to extract structured information from Wikipedia. The dataset uses the DBPedia ontology  to provide hierarchical labels for a large collection of Wikipedia articles.

We divide \gls{WOS} and \gls{MN-DS} into 80\% training and 20\% test, while keeping the published data split for DBPedia ($\sim$18\% test, $\sim$10\% validation). We do not use the validation set.

\begin{table}[t!]
\begin{scriptsize}
\begin{center}
\begin{tabular}{p{0.45\linewidth}p{0.078\linewidth}p{0.13\linewidth}p{0.12\linewidth}}
\toprule
\multirow{2}{*}{\textbf{Article}}& \multicolumn{3}{c}{\textbf{Taxonomic Labels}}\\
\cmidrule{2-4}
& Lvl. 1 & Lvl. 2 & Lvl. 3\\
\midrule
William Alexander Massey (October 7, 1856 – March 5, 1914) $\dots$ & \multirow{3}{*}{Agent} & \multirow{3}{*}{Politician} & \multirow{3}{*}{Senator} \\
\midrule
Lions is the sixth studio album by American rock band The $\dots$ & \multirow{2}{*}{Work} & \multirow{2}{*}{MusicalWork} & \multirow{2}{*}{Album} \\
\midrule
Pirqa (Aymara and Quechua for wall, hispanicized spelling  $\dots$ & \multirow{2}{*}{Place} & \multirow{2}{*}{NaturalPlace} & \multirow{2}{*}{Mountain} \\
\bottomrule
\end{tabular}
\end{center}
\end{scriptsize}
\caption{Examples from the DBP dataset. 
%Wikipedia articles must be classified into hierarchical labels.
}
\label{tab:dbp_example}
\end{table}
\setlength\arrayrulewidth{1pt}
\begin{table}[t]
    \begin{center}\resizebox{1\linewidth}{!}{
    \footnotesize
    \begin{tabular}{lrrrr}
        \toprule
         \textbf{Dataset} & \# Samples & Lvl. 1 & Lvl. 2 & Lvl. 3\\ 
         \midrule
        Web Of Science (WOS) & 46,985 & 7 & 134 & - \\
        MN-DS News (MN-DS) & 10,917 & 17 & 109 & -  \\
        DBPedia (DBP) & 342,782 & 9 & 70 & 219 \\
        \bottomrule
    \end{tabular}}\end{center}
  \caption{Hierarchical classification dataset statistics 
  %used for the downstream evaluation. 
  The levels (Lvl. 1-3) indicate the number of labels per taxonomic depth, where 0 indicates the root.}
  \label{tab:downstream_stats} 
\end{table}

\paragraph{Classification Approach}
We use logistic regression classifiers on top of sentence-transformer embeddings (\texttt{MiniLM-L6}) of the article texts. 
%For hierarchical classification, 
We perform  standard top-down prediction by fitting one classifier per parent node \cite{electronics13071199}. At each level, one of the children of the previously selected node is predicted in a multi-class setting. This results in a number of classifiers equal to the number of inner %concepts 
nodes
%of the taxonomy 
plus the root. 

\paragraph{Setup}
Following the intrinsic evaluation experimental setup, we degrade versions of all label taxonomies (WOS, DBP, MN-DS) by randomly relocating sub-graphs of the original label taxonomies.\ We again mutate between 2\% and 32\% of the taxonomy nodes and calculate the downstream F1 score by fitting the hierarchical classification outlined above using the degraded label taxonomy. As before, reference-free metric scores are calculated every $2^x$ mutations 
%(5 times) 
with $x = 1, \dots, 5$. We sample 100 different degraded taxonomies for each dataset.
% resulting in 500 degraded versions.\ 
The correlation is then calculated between the macro F1 score on the lowest-level taxonomic labels and the reference-free metrics.
The assumption is therefore that the classifier performance  is associated with the quality of the label taxonomy.
We use the macro F1, since the datasets are not balanced and we want to give equal importance to all classes.

\paragraph{Results}

%\begin{figure}[b!]
\begin{figure}
\centering
\includegraphics[width=1.0\linewidth]{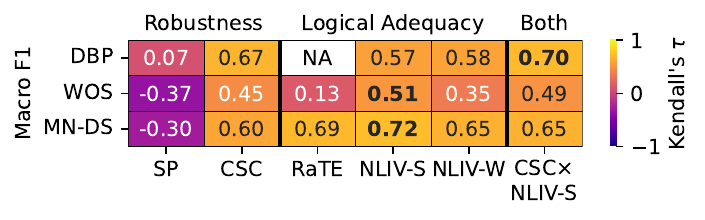}\\
\caption{Correlations between macro F1 on the downstream task and our reference-free metrics. All correlations are statistically significant with $\alpha = 0.001$.}
\label{fig:corrs_downstream}
\end{figure}

% \begin{figure*}[t!]
% \centering
% \includegraphics[width=0.90\linewidth]{figures/f1_vs_metrics_datasets_WOS.pdf}
% \caption{Scatterplot of reference-free metrics and downstream performances during the extrinsic evaluation on the WOS dataset. The curve indicates the best fit of an exponential model of form $y = a \cdot e^{bx} + c$.} 
% \label{fig:metric_surfaces_downstream}
% \end{figure*}

Fig.~\ref{fig:corrs_downstream} shows the correlations between macro F1 on the downstream task and reference-free metrics. All versions of our metrics show substantial associations with the downstream performance, whereas RaTE and SP do not, or do so only partially. On DBP, RaTE never predicts the true parent of a node in the top-$k$, resulting in a constant score and therefore an undefined correlation.\
This is one of the reasons \citet{langlais-gao-2023-rate} perform domain-specific fine-tuning of masked language models.\
SP is even negatively associated with downstream performance on WOS and MN-DS, likely because of the high branching factor of the ground truth label taxonomy, which makes intruders in sibling groups very likely.

For an intuition on how the metrics behave, we show scatter plots and a fitted exponential on the space of the collected metric scores of taxonomies mutated during the experiments  (Fig.~\ref{fig:scatter_downstream_all} in App. \ref{app:sec:extended:scatter}). CSC and NLIV can, to an extent, predict downstream F1, as indicated by the $R^2$. While RaTE shows a positive tendency, there is higher variance and it does not work for DBP without tuning as shown 
%previously 
in Fig. \ref{fig:corrs_downstream}. The scatter plot for SP clearly shows its negative association with 
%downstream 
F1. 
\subsection{Extended Analysis}
\begin{figure}[H]
\centering
\includegraphics[width=1.0\linewidth]{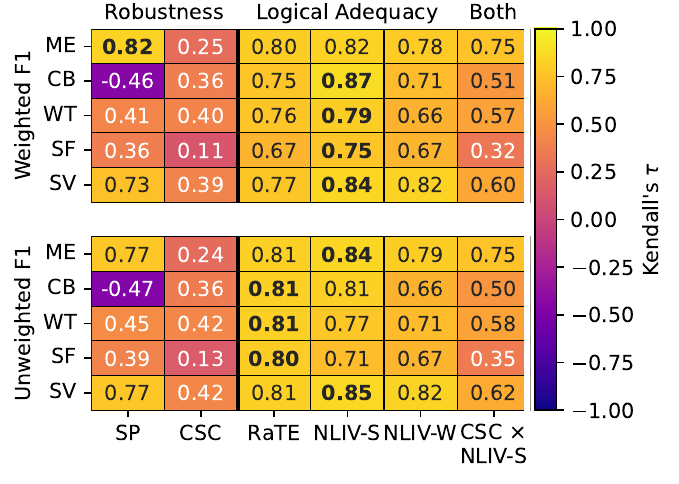}\\
\caption{Correlations between F1 and reference-free metrics when moving a node in its vicinity (weighted by WPS). All correlations are significant with $\alpha = 0.001$. 
%We calculated F1 scores with and without class weights, assigning weights based on the number of descendants of a concept.
}
\label{fig:corrs_wps}
\end{figure}

% Subsequently, we investigate whether the underlying assumptions of modelling decisions hold, specifically by looking at the performance of NLI for discriminating parent-child edges. To test the reliability of our metrics, we subject them to variations in input and models.
We carry out further analysis of underlying assumptions and modelling decisions by subjecting our metrics to variations in input and base models. 
%For the former, we examine NLI performance in discriminating parent-child edges. 
%We examine the effect of variations in inputs and base models on our metrics.
\begin{figure}[t!]
\centering
\includegraphics[width=1.0\linewidth]{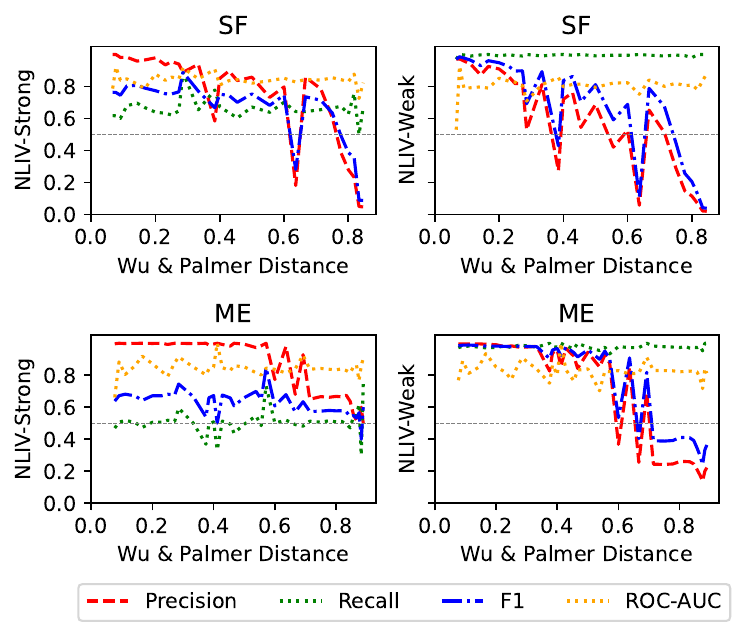}\\
\caption{Binary classification performance 
%on the identification of true or false edges 
with zero-shot NLI 
%(weak and strong) 
on the 
%SemEval-Food 
SF 
and 
%MeSH
ME 
taxonomies.  The x-axis is the WPS
%Wu \& Palmer distance (1 - Wu \& Palmer Similarity) 
between the two concepts in the ground truth taxonomy. The gray dashed line is the random baseline. 
%expected performance of a random guess.
}
\label{fig:nli_val_small}
\end{figure}
\paragraph{Realistic Errors}
In the experiments in Section~\ref{sec:intrinsic}, we degrade taxonomies by moving  either \textit{non-leaves} or \textit{all} concepts. This has the  advantage of making no assumptions about the error modes of potential generation or completion algorithms. However, since errors in taxonomy generation or completion methods often place nodes under similar but wrong hypernyms, we conduct an additional intrinsic evaluation, where a concept's sub-graph can only be moved under one of the top 100 closest concepts according to
WPS\footnote{WPS is also used as sampling weight.}.
%the Wu \& Palmer similarity. 
% Note that the previous experiments, where we move \textit{all} and \textit{non-leaf} concepts, are still important, since they do not introduce assumptions about the error modes of potential generation or completion algorithms in contrast to this experiment. 
The results are shown in Fig. \ref{fig:corrs_wps}. The correlations for the NLIV metrics remain high but are lower for \gls{CSC}, which is expected. \gls{CSC}, by definition, focusses on the global shape of the taxonomy, and moving concepts locally will only have a minor effect on it.

% Correlations

\paragraph{Approximating Parent-Child Likelihoods}
\label{sec:nli_validation}
Since our logical adequacy metric depends on NLI to estimate the probability that a parent-child edge is adequate, %$P(A | \langle c_i, c_{i+1} \rangle)$, 
it would be undermined by low NLI accuracy. We therefore define a binary classification task to evaluate the NLI model's suitability. 
% To this end, 
We use existing edges as positive  instances and an equal amount of random, unrelated concept pairs as negatives.  We consider a strong (NLIV-S, Eq. \ref{eq:nliv_strong})  and weak (NLIV-W, Eq. \ref{eq:nliv_weak}) version. 
% In case NLI is inaccurate in this setting, our logical adequacy metrics become meaningless.
Fig.~\ref{fig:nli_val_small} shows the NLI binary classification performance on SemEval-Food (SF) and MeSH (ME). See Tab.~\ref{tab:approx_perf} in App.~\ref{app:sec:extended} for the full set of scores\footnote{We also tried using concept names instead of descriptions and found that performance is almost entirely conserved.}. 
As expected, we observe higher precision compared to recall with NLIV-S, while the opposite is true for NLIV-W. 
The ROC-AUC hovering around 0.8 for both datasets and metrics indicates that the NLI models are indeed quite discriminative and thus applicable to our metrics. 
% Precision, recall, and F1 
Scores change as the two concepts of a pair become more unrelated, implying that the decision boundary depends on the taxonomic distance of a pair.

\begin{figure}[t!]
\centering
\includegraphics[width=0.90\linewidth]{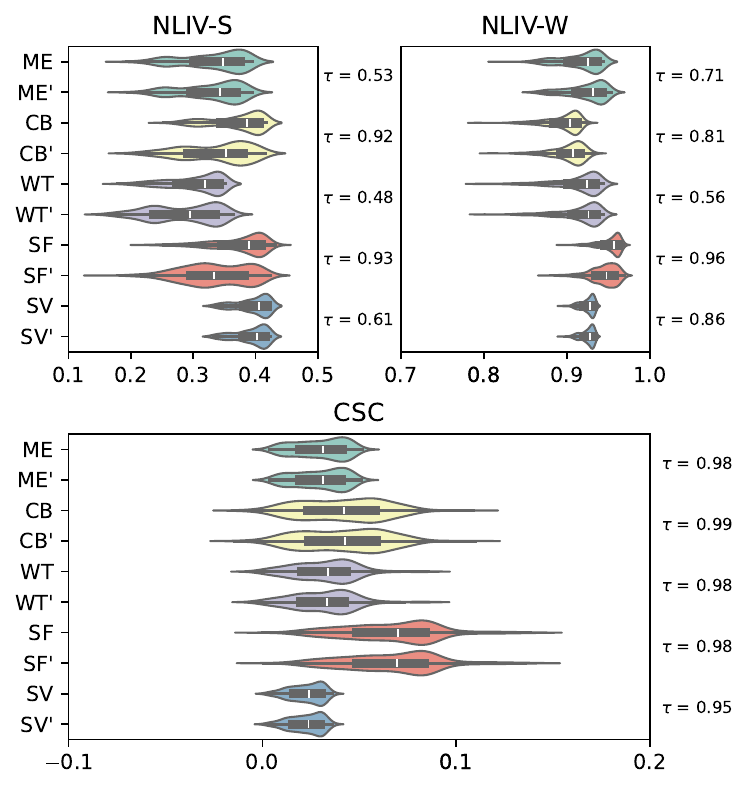}
  \caption{Violin plots showing the distribution of our scores when replacing concept names with synonyms using WordNet. The right hand side shows the Kendall's $\tau$ between the two distributions as a measure of their monotonic association. The distributions of perturbed taxonomies are indicated with primes (e.g., CB$^\prime$).}
  \label{fig:perturbations-violins}\bigskip

% pushes the next subfigure to the bottom of the minipage

  \includegraphics[width=0.90\linewidth]{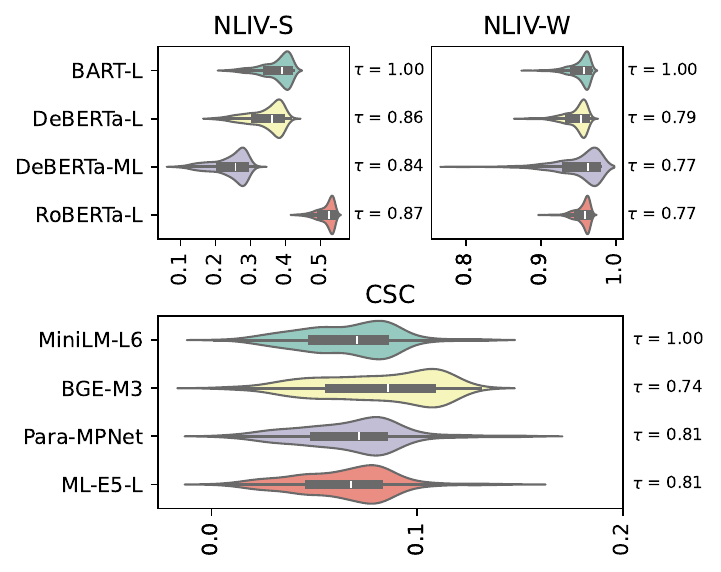}
  \caption{Violin plots showing the distribution of our scores when changing backbone models on SemEval-Food. The righthand side shows the Kendall's $\tau$ between the model score distribution and the models used in our original experiments (BART-L and MiniLM-L6).}
  \label{fig:model-violins}\medskip

\end{figure}

\paragraph{Sensitivity to Input Perturbations}
To test how stable our metrics are to different, semantically identical taxonomies, we reuse our intrinsic evaluation setup,
but for each mutated taxonomy, we create a corresponding taxonomy that has between 0\% and 100\% 
%(uniformly sampled percentage) 
of its concepts replaced by synonyms.
%To this end, 
We map concept names to WordNet \cite{Fellbaum2019-dh} where possible and sample a 
synonym from the synset of the most common meaning. We then calculate scores for both 
taxonomies and compare the rank correlations using Kendall's $\tau$. If our metrics are sufficiently robust to
the induced perturbations, we expect large $\tau$'s between mutated taxonomies and their perturbed versions. The results are shown in 
Fig.~\ref{fig:perturbations-violins}.
% shows the score distributions for taxonomies perturbed by synonym replacement  vs. the original mutation. 
While \gls{CSC} is robust with high Kendall's $\tau$ between the two distributions, 
%per dataset, 
the NLIV metrics are more sensitive to changes in inputs, which aligns with the literature on NLI models \cite{arakelyan-etal-2024-semantic}. However, NLIV metrics still show relatively high association between the original and perturbed taxonomies when compared to RaTE (see plot for RaTE and SP in Fig.~\ref{fig:perturbations-violins-full}, App.~\ref{app:sec:extended}). The stability of NLIV also seems to be dataset-dependent, as indicated by the 
%large 
differences in 
%Kendall's 
$\tau$ for the same metric.

\paragraph{Sensitivity to Model Changes}
To test how robust our methodology is to changes in model, we evaluated four different sentence transformers (MiniLM-L6, BGE-M3, ML-E5-L, and Para-MPNet)
and NLI-models (RoBERTa-L, DeBERTa-L, BART-L, DeBERTa-ML) against the same taxonomy mutations and compared the ranking of the taxonomies using Kendall's $\tau$.
%\footnote{See App.~\ref{app:model_ids} 
%for the full model identifiers.} 
The results are shown in Fig.~\ref{fig:model-violins}.
Scores cannot be directly compared between models due to the shifting distributions, but 
the consistently high $\tau$ 
% indicates that the rankings of the  taxonomies are highly associated between models, meaning 
suggests
that the same taxonomies can be compared using different models and the ranking will stay fairly consistent.

\clearpage\section{Conclusion}
We propose two novel 
%intrinsic 
metrics for assessing taxonomy quality, CSC and NLIV, that offer greater resilience across 
%different 
domains and mutations than existing ones without requiring fine-tuning.\ We show that 
%the two metrics 
they
are highly associated with taxonomy quality as measured by comparison to %ground truth 
gold
taxonomies, 
%They are also 
and are 
predictive of hierarchical classification performance when applied to label taxonomies. 
 We envisage using these metrics primarily to compare alternative taxonomies, 
 %of the same domain (produced by different models). %However, 
 %it may also be possible to use these metrics 
 but they could also be used
 to pinpoint individual errors
 by, for example, 
 %One could look at 
 using classification probabilities predicted by NLIV to identify very unlikely classification walks, or  examining concept pairs where semantic similarity and taxonomic similarity show the highest disagreement. 
Further research is needed to explore these possibilities, as well as to investigate how these metrics could be
%be combined and 
leveraged as an optimization objective or how they relate to human intuition about taxonomy quality.

%\clearpage
\section{Limitations}

\begin{itemize}
\item While our proposed metrics are designed to be domain independent, the pre-trained models we use are not. 
As a result, the computed scores may reflect biases inherent in the models used for semantic similarity and \gls{NLI}.
\item Our assessment of logical adequacy %heavily based on
depends on
\gls{NLI}. Before applying our method, 
we recommend evaluating the model performance in the taxonomy domain through experiments similar to those in Section~\ref{sec:nli_validation}.
\item While we observe in experiments that the metrics can differentiate taxonomy quality within the same concept space, 
we do not expect our metrics to be directly comparable between different concept spaces due to the varying performances
of baseline models in these spaces and the potentially differing goals and properties of their taxonomies.
\end{itemize}

\section{Acknowledgements}
We would like to express our sincere appreciation to Betty Bossi\footnote{\url{https://www.bettybossi.ch/}} for their support of this research project and for providing us with their taxonomy used for recipe development. This research is supported through computing resources by the ADAPT
Centre for Digital Content Technology, which is funded under the SFI Research Centres Programme (Grant 13/RC/2106\_P2) and is co-funded under the European Regional Development Fund (ERDF).
The authors thank the reviewers for their insightful and helpful comments. One author was supported in part by the Swiss National Science Foundation (SNSF) under grant 20HW-1\_228541.

\bibliography{refs}

\clearpage
\appendix
\section{Quality Criteria}
\label{app:sec:criteria}
The consolidated quality criteria of taxonomies according to \citet{https://doi.org/10.1111/exsy.13098} are detailed below.
\begin{itemize}[itemsep=0.0em]
\item \textbf{Comprehensiveness}: the capacity to categorize all objects within the taxonomy's target domain.
\item \textbf{Robustness}: how well a taxonomy separates concepts, depending on orthogonality (concepts represent distinct ideas) and cohesiveness (sibling concepts are closely related).
\item \textbf{Conciseness}: the capacity to classify objects using a minimal number of concepts; an external quality observed in practical application.
\item \textbf{Extensibility}: the capability to accommodate structural modifications such as adding, altering, or removing roots or concepts.
\item \textbf{Explanatory}: allows users to classify objects by their characteristics or infer characteristics from placement.
\item \textbf{Mutual exclusiveness}: uniquely identifies an object, ensuring no object falls under different concepts within the same dimension; an external quality observed during application.
\item \textbf{Reliability}: ensures consistent classification across different coders.
\end{itemize}

\section{Wu \& Palmer Similarity}
\begin{figure}[!hb]
\centering
\includegraphics[width=1.0\linewidth]{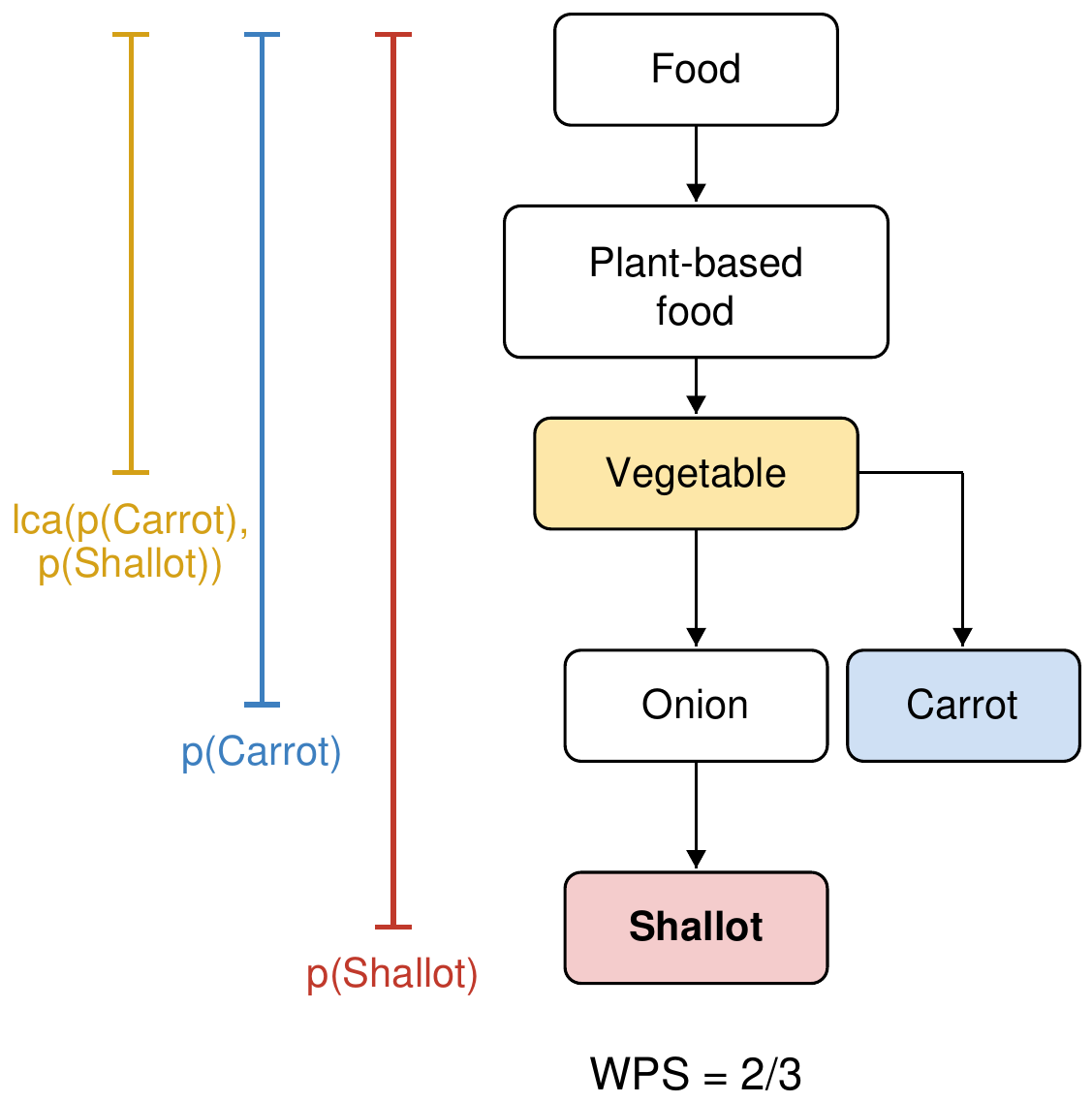}
\caption{Example of \gls{WPS} calculation.}
\label{fig:wps}
\end{figure}

Fig.~\ref{fig:wps} illustrates how to calculate the \gls{WPS} (Eq.~\ref{eq:wpsapp}) using an example.

\begin{small}
\begin{eqnarray}
\text{W}_{c_a c_b} &=& \frac{2 \cdot \mathtt{lca}(p(c_a), p(c_b))}{\lvert p(c_a)\rvert + \lvert p(c_b)\rvert}
\label{eq:wpsapp}
\end{eqnarray}
\end{small}

\section{Logical Adequacy}
\label{app:log_ad}

The joint probability (Eq. \ref{eq:jointp}) becomes the product of edge probabilities due to the assumption that edge probabilities are independent.

\begin{small}
\begin{equation}
P(A|C) := \prod_{i=0}^{k-1} P(A | \langle c_i, c_{i+1} \rangle)
\label{eq:jointp}
\end{equation}
\end{small}

Our inspiration for normalising the probability by length (Eq.~\ref{eq:pnorm2}) using the geometric mean stems from the well known perplexity measure \cite{10.1121/1.2016299}.

\begin{small}
\begin{equation}
\tilde{P}(A|C) := \left[ \prod_{i=0}^{k-1} P(A | \langle c_i, c_{i+1} \rangle)\right]^{\frac{1}{k}}
\label{eq:pnorm2}
\end{equation}
\end{small}

Since we have no notion of what the distribution of classifications looks like during the application of the taxonomy, we define it as uniform to avoid favouring any specific use case. This means $P(C)$ of  Eq. \ref{eq:unionapp} becomes $\frac{1}{|\V|}$ for all $C$.

\begin{small}
\begin{equation}
\begin{split}
\tilde{P}(A) &:= \sum\limits_{C} \tilde{P}(A | C) \times P(C)
= \frac{1}{|\V|} \sum\limits_{C} \tilde{P}(A | C)\\
\end{split}
\label{eq:unionapp}
\end{equation}
\end{small}

\section{Hearst Patterns}
\label{app:hearst}
Here we show the full list of pattern instances used for the approximation of edge adequacy in the NLIV metrics.

\begin{small}
\begin{itemize}[leftmargin=*]
\itemsep0em 
\vspace*{-0.3em}\item \textbf{Pattern (1):} \label{eq:pattern1_app}
\vspace*{-0.3em}\begin{align*}
&s(\langle c_i, c_{i+1} \rangle) \in S, \; S = \{\\
&\quad ''D(c_{i+1}). A(c_{i+1}) \, N(c_{i+1}) \text{ is a type of } N(c_i)'', \\
&\quad ''D(c_{i+1}). A(c_{i+1}) \, N(c_{i+1}) \text{ is an example of } N(c_i)'', \\
&\quad ''D(c_{i+1}). A(c_{i+1}) \, N(c_{i+1}) \text{ is } A(c_i) \, N(c_i)'', \\
&\quad ''D(c_{i+1}). A(c_{i+1}) \, N(c_{i+1}) \text{ is a kind of } N(c_i)''\}
\end{align*}

\vspace*{-0.3em}\item \textbf{Pattern (2):} \label{eq:pattern2_app}
\vspace*{-0.3em}\begin{align*}
&s(\langle c_i, c_{i+1} \rangle) \in S, \; S = \{\\
&\quad ''D(c_{i+1}). A(c_i) \, N(c_i) \text{ such as } A(c_{i+1}) \, N(c_{i+1})'', \\
&\quad ''D(c_{i+1}). \text{Such } N(c_i)_{\text{plural}} \text{ as } N(c_{i+1})''\}
\end{align*}

\vspace*{-0.3em}\item \textbf{Pattern (3):} \label{eq:pattern3_app}
\vspace*{-0.3em}\begin{align*}
&s(\langle c_i, c_{i+1} \rangle) = \\
&\quad ''D(c_{i+1}). A(c_{i+1}) \, N(c_{i+1}) \text{ or other } N(c_i)_{\text{plural}}''
\end{align*}

\vspace*{-0.3em}\item \textbf{Pattern (4):} \label{eq:pattern4_app}
\vspace*{-0.3em}\begin{align*}
&s(\langle c_i, c_{i+1} \rangle) = \\
&\quad ''D(c_{i+1}). A(c_{i+1}) \, N(c_{i+1}) \text{ and other } N(c_i)_{\text{plural}}''
\end{align*}

\vspace*{-0.3em}\item \textbf{Pattern (5):} \label{eq:pattern5_app}
\vspace*{-0.3em}\begin{align*}
&s(\langle c_i, c_{i+1} \rangle) = \\
&\quad ''D(c_{i+1}). N(c_i)_{\text{plural}}, \text{ including } N(c_{i+1})''
\end{align*}

\vspace*{-0.3em}\item \textbf{Pattern (6):} \label{eq:pattern6_app}
\vspace*{-0.3em}\begin{align*}
&s(\langle c_i, c_{i+1} \rangle) = \\
&\quad ''D(c_{i+1}). N(c_i)_{\text{plural}}, \text{ especially } N(c_{i+1})''
\end{align*}

\end{itemize}
\end{small}

\section{Semantic Proximity}
\label{app:sec:semprox}

Eq.~\ref{eq:semprox} shows the \gls{SP} where \(\mathcal{V}\) is the set of all concepts, $\mathcal{L}$ denotes the set of all leaf concepts in $\mathcal{V}$, \(\mathcal{G}(l)\) is the set of sibling leaves of the leaf $l$, \(i\) and $j$ represent leaves, and \(\mathtt{sim}(i, j)\) denotes the similarity measure between the two. The indicator function \(\mathbb{I}(\cdot)\) evaluates to 1 if the expression inside is true and to 0 if it is false. This formulation, although different in notation, is equivalent to the original formulation in the work of \citet{https://doi.org/10.1111/exsy.13098}, but expressed in terms of sets.

\begin{scriptsize}
\begin{align}
%\mathcal{L} &= \{v \in \mathcal{V}\; | \; (\not \exists e)[e \in \mathcal{E} \wedge c \in{\mathcal{V}} \wedge e =(v, c)]  \}\\
SP &= \frac{1}{|\mathcal{L}|} \sum_{l \in \mathcal{L}} 1- \mathbb{I} \left( \min_{i, j \in \mathcal{G}(l), i \neq j} \mathtt{sim}(i, j) > \!\min_{k \in \mathcal{V} \setminus \mathcal{G}(l)} \mathtt{sim}(l, k) \right)
\label{eq:semprox}
\end{align}
\end{scriptsize}

\section{Experiment Details}

\subsection{Degrading Taxonomies Randomly}
We sample two unrelated concepts (neither are descendants or ancestors of each other) in the taxonomy and change the parent of the first concept to the second concept. In some taxonomies, concepts can have multiple parents. In such cases, we remove all current parents before adding the concept at the new parent. Tab.~\ref{tab:randomizations} shows the number of sampled degraded taxonomies per taxonomy. For the experiments in the extended analysis, Tab.~\ref{tab:randomizations} holds as well. The extrinsic evaluation features 100 degradations for each dataset; therefore, there are a total of 500 samples (5 for each degradation).

\begin{table*}[b]
\begin{normalsize}
\centering
\resizebox{0.8\linewidth}{!}{
\begin{tabular}{lrrrrr}
\toprule
\multirow{2}{*}{\textbf{Dataset}} & \multicolumn{2}{c}{\textbf{All}} && \multicolumn{2}{c}{\textbf{Non-Leaf}}\\
\cmidrule{2-3}
\cmidrule{5-6}
& Number samples & Number degradations && Number samples & Number degradations \\
\midrule
MeSH & 250 & 50  && 250 & 50 \\
CookBook & 500 & 100 && 500 & 100 \\
WikiTax & 500 & 100 && 500 & 100 \\
SemEval-Food & 500 & 100 && 500 & 100 \\
SemEval-Verb & 250 & 50 && 250 & 50 \\
\bottomrule
\end{tabular}}
\caption{The number of samples and degradations per taxonomy for the empirical validation.}
\label{tab:randomizations}
\end{normalsize}\bigskip\bigskip\bigskip\bigskip

\begin{normalsize}
\begin{center}
\resizebox{0.8\linewidth}{!}{
\begin{tabular}{l | l | l}
\toprule
\textbf{Model Name} & \textbf{Huggingface Link} & \textbf{Publication} \\
\midrule
BART-L & \href{https://huggingface.co/facebook/bart-large-mnli}{facebook/bart-large-mnli} & \citet{lewis-etal-2020-bart} \\
BGE-M3 & \href{https://huggingface.co/BAAI/bge-m3}{BAAI/bge-m3} & \citet{bge-m3} \\
DeBERTa-L & \href{https://huggingface.co/microsoft/deberta-large-mnli}{microsoft/deberta-large-mnli} & \citet{he2021deberta} \\
DeBERTa-ML & \href{https://huggingface.co/MoritzLaurer/DeBERTa-v3-large-mnli-fever-anli-ling-wanli}{MoritzLaurer/DeBERTa-v3-large-mnli-fever-anli-ling-wanli} & \citet{he2021deberta,laurer2022lessannotating} \\
MiniLM-L6 & \href{https://huggingface.co/sentence-transformers/all-MiniLM-L6-v2}{sentence-transformers/all-MiniLM-L6-v2} & \citet{reimers-2019-sentence-bert} \\
ML-E5-L & \href{https://huggingface.co/intfloat/multilingual-e5-large-instruct}{intfloat/multilingual-e5-large-instruct} & \citet{wang2024multilingual} \\
Para-MPNet & \href{https://huggingface.co/sentence-transformers/paraphrase-multilingual-mpnet-base-v2}{sentence-transformers/paraphrase-multilingual-mpnet-base-v2} & \citet{reimers-2019-sentence-bert} \\
RoBERTa-L & \href{https://huggingface.co/FacebookAI/roberta-large-mnli}{FacebookAI/roberta-large-mnli} & \citet{liu2019roberta} \\
\bottomrule
\end{tabular}}
\caption{Huggingface identifiers and links for model abbreviations along with their corresponding publications.}
\label{tab:model_ids}
\end{center}
\end{normalsize}\bigskip\bigskip\bigskip\bigskip

\centering
\resizebox{0.8\linewidth}{!}{
\begin{normalsize}\begin{tabular}{lrrrrrrrrr}
\toprule
\multirow{2}{*}{\textbf{Dataset}} & \multicolumn{4}{c}{\textbf{NLIV-Strong}} && \multicolumn{4}{c}{\textbf{NLIV-Weak}}\\

\cmidrule{2-5}
\cmidrule{7-10}

& ROC-AUC & Precision & Recall & F1 && ROC-AUC & Precision & Recall & F1\\
\midrule
\multicolumn{10}{c}{With concept descriptions} \\
\midrule
MeSH & 0.84 & 0.85 & 0.50 & 0.63 && 0.82 & 0.48 & 0.98 & 0.65 \\
CookBook & 0.90 & 0.92 & 0.68 & 0.78  && 0.87 & 0.58 & 0.95 & 0.72 \\
WikiTax & 0.76 & 0.78 & 0.59 & 0.67  && 0.78 & 0.57 & 0.92 & 0.70 \\
SemEval-Food & 0.84 & 0.79 & 0.65 & 0.72 && 0.81 & 0.53 & 0.99 & 0.69 \\
SemEval-Verb & 0.70 & 0.56 & 0.67 & 0.61 && 0.66 & 0.43 & 0.99 & 0.60 \\
\midrule
\multicolumn{10}{c}{With concept names (no descriptions)} \\
\midrule
SemEval-Food & 0.84 & 0.79 & 0.65 & 0.72 && 0.81 & 0.53 & 0.99 & 0.69 \\
MeSH & 0.84 & 0.84 & 0.52 & 0.64 && 0.84 & 0.47 & 1.00 & 0.64 \\
\bottomrule
\end{tabular}\end{normalsize}}
\caption{Performance on the binary classification of true or false parent-child edges. The lower table shows experiments in simply using concept names instead of descriptions. Performance is preserved in cases where concept descriptions are unavailable.}
\label{tab:approx_perf}
\end{table*}

\subsection{Model Identifiers}
\label{app:model_ids}
Tab.~\ref{tab:model_ids} shows the corresponding HuggingFace repositories for the models used in the experiment.

\subsection{WikiData Taxonomy}
\label{app:sec:wikitax}
We extract the WikiTax taxonomy using the edges \textit{subclass of}, \textit{instances of} and \textit{subproperty of} (Wikidata identifiers P279, P31 and P1647)\footnote{\url{https://github.com/nichtich/wikidata-taxonomy}}. 

\subsection{Correlation Coefficients}
We use rank correlation since we do not expect a linear relationship between metric scores, as we expect their rankings to align rather than their absolute values.

\section{Extended Results}\label{app:sec:extended}
\subsection{Extrinsic Evaluation}\label{app:sec:extended:scatter}
Fig.~\ref{fig:scatter_downstream_all} shows scatter plots and a fitted exponential on the space of the collected metric scores of taxonomies mutated during the experiments. CSC and NLIV can, to an extent, predict downstream F1, as indicated by the $R^2$. While RaTE shows a positive tendency, there is higher variance and it does not work for DBP without tuning as shown 
%previously 
in Fig. \ref{fig:corrs_downstream}. The scatter plot for SP clearly shows its negative association with 
%downstream 
F1. 

\subsection{Stability of Metrics}
\begin{figure*}[t]
\centering
\includegraphics[width=0.65\linewidth]{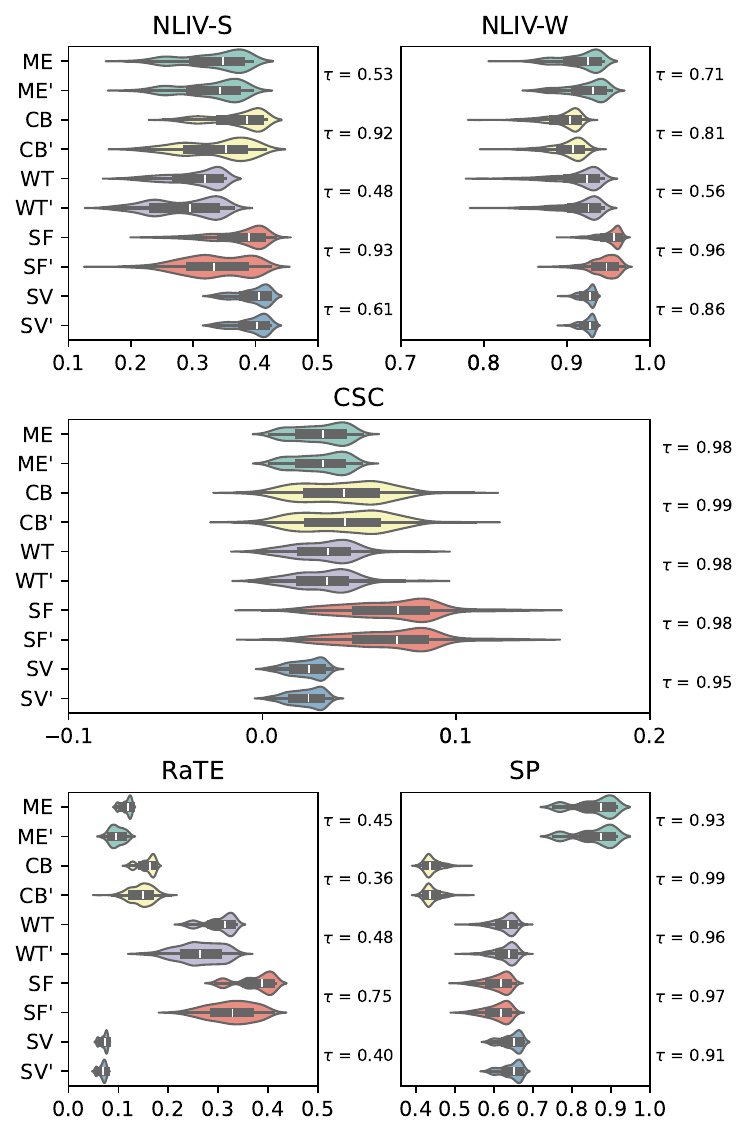}
  \caption{Violin plots showing the distribution of our scores when replacing concept names with synonyms using WordNet. This is the full plot including RaTE and SP metrics. The right hand side shows the Kendall's $\tau$ between the two distributions as a measure of their monotonic association. The distributions of perturbed taxonomies are indicated with primes (e.g., CB$^\prime$).}
  \label{fig:perturbations-violins-full}\bigskip
\end{figure*}

Fig.~\ref{fig:perturbations-violins-full} shows the violin plots for perturbed vs. mutated taxonomies now with RaTE and SP included.

\subsection{Metrics}
\begin{figure*}[!t]
\centering
\subfloat[Our metrics]{
\includegraphics[width=0.9\textwidth] {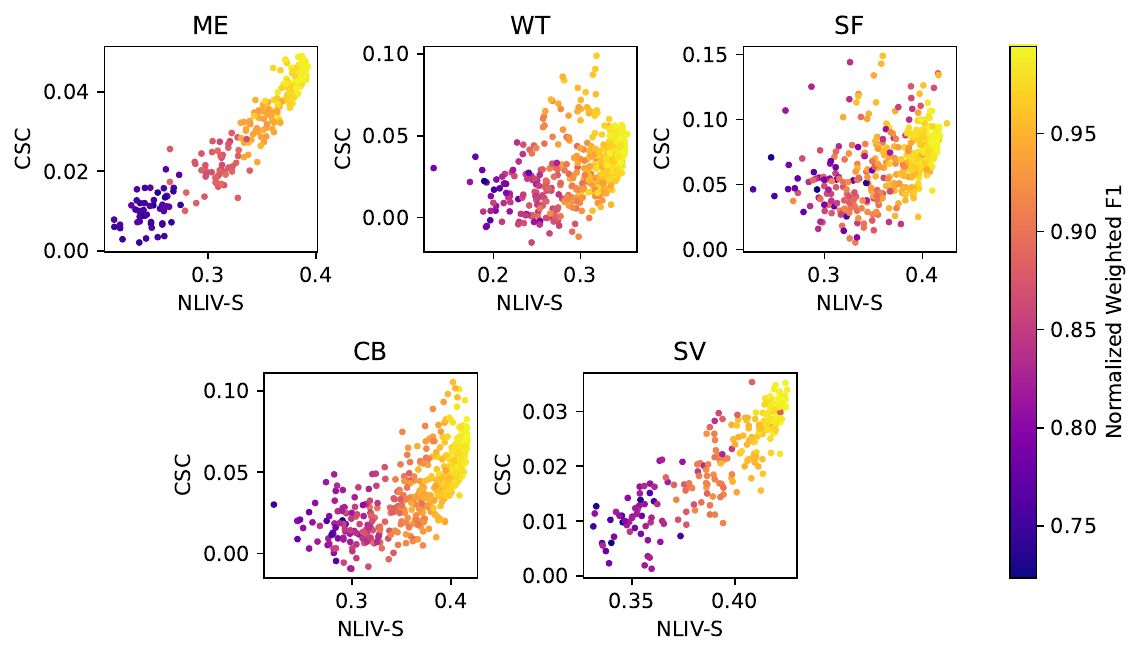}}\\
\subfloat[Existing metrics]{
\includegraphics[width=0.9\textwidth] {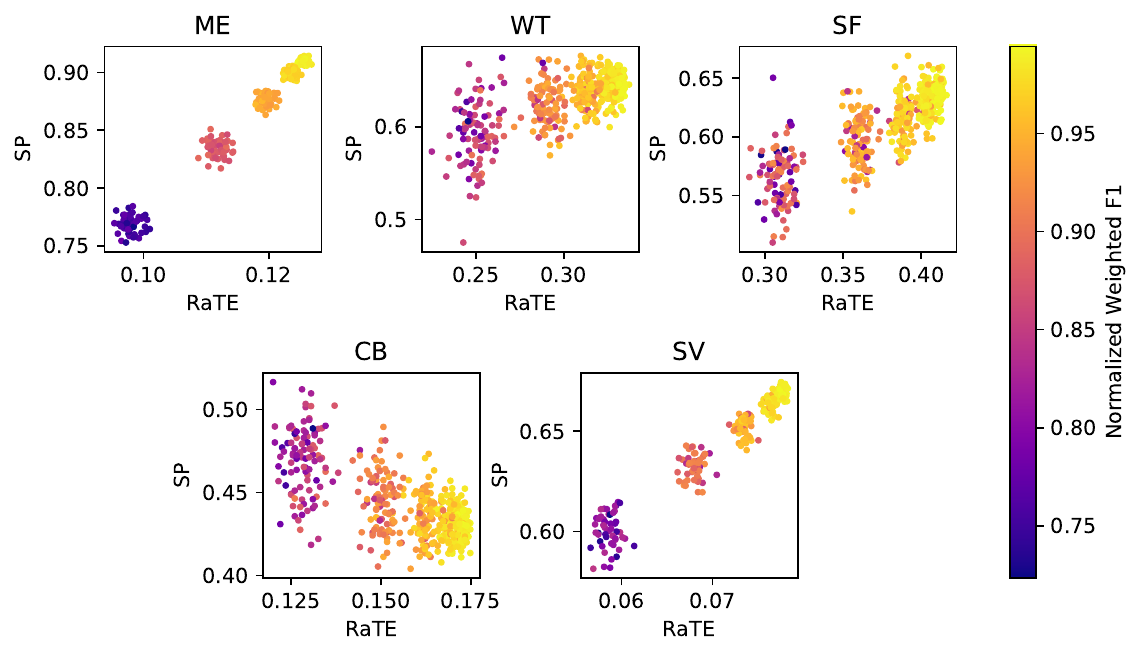}}
\caption{Scatter plots for the intrinsic evaluation samples with the color indicating the weighted F1 score against the ground truth taxonomy.}
\label{fig:scatter_all_weighted}
\end{figure*}

\begin{figure*}[!b]
\centering
\subfloat[Our metrics]{
\includegraphics[width=0.9\textwidth] {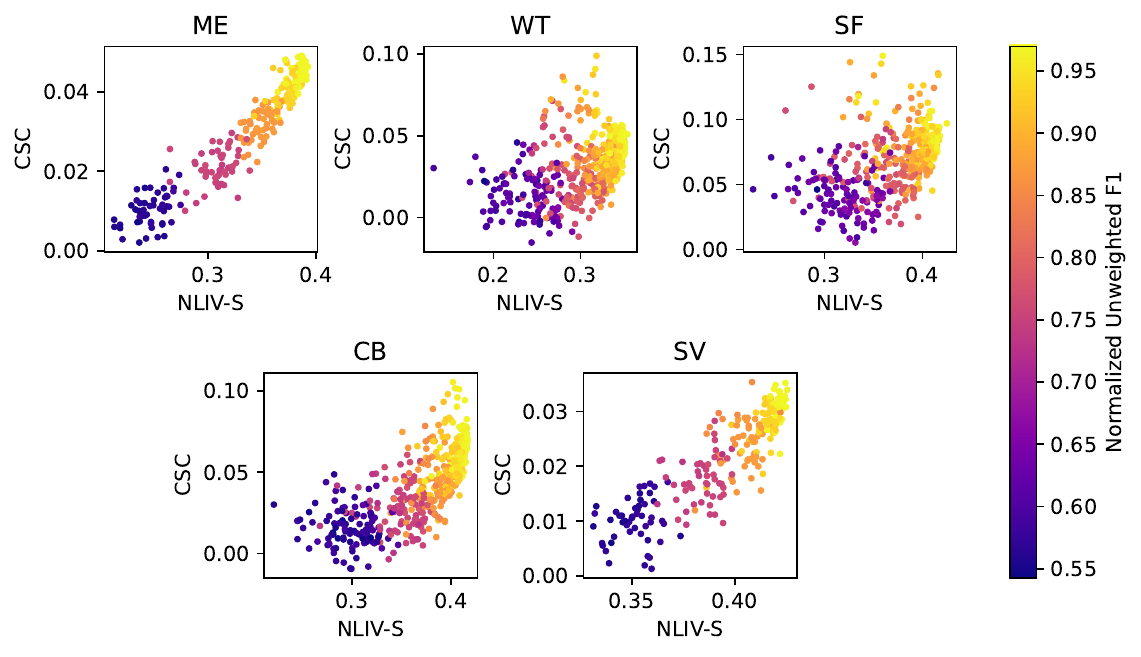}}\\
\subfloat[Existing metrics]{
\includegraphics[width=0.9\textwidth] {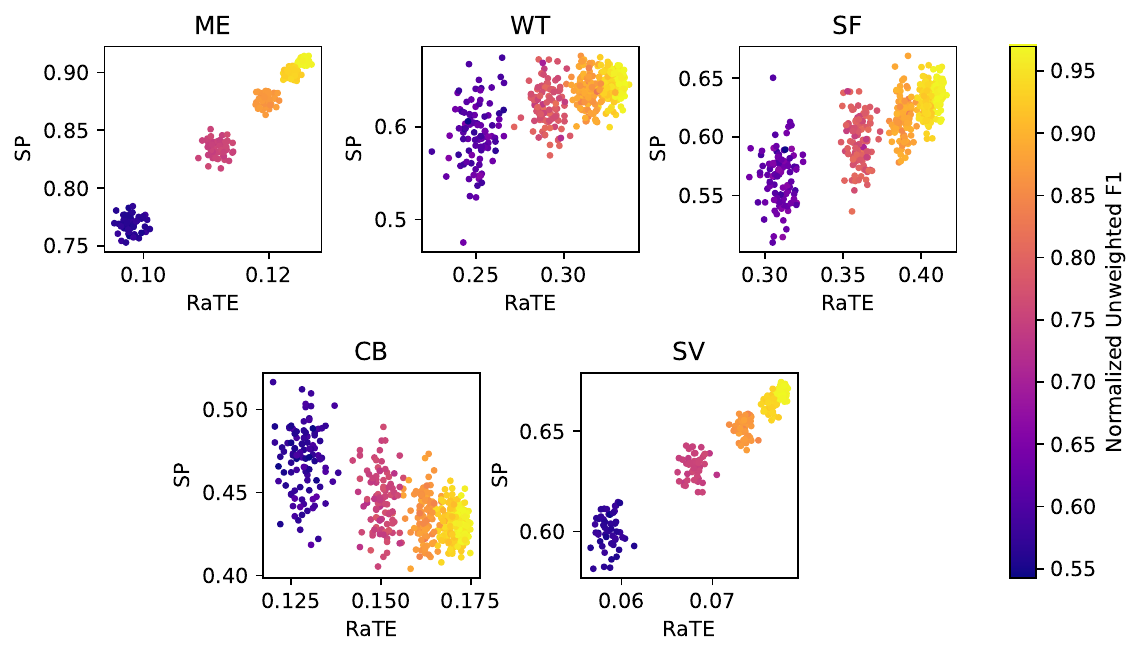}}
\caption{Scatter plots for the intrinsic evaluation samples with the color indicating the unweighted F1 score against the ground truth taxonomy.}
\label{fig:scatter_all_unweighted}
\end{figure*}

\begin{figure*}[!b]
\centering
\includegraphics[width=\textwidth] {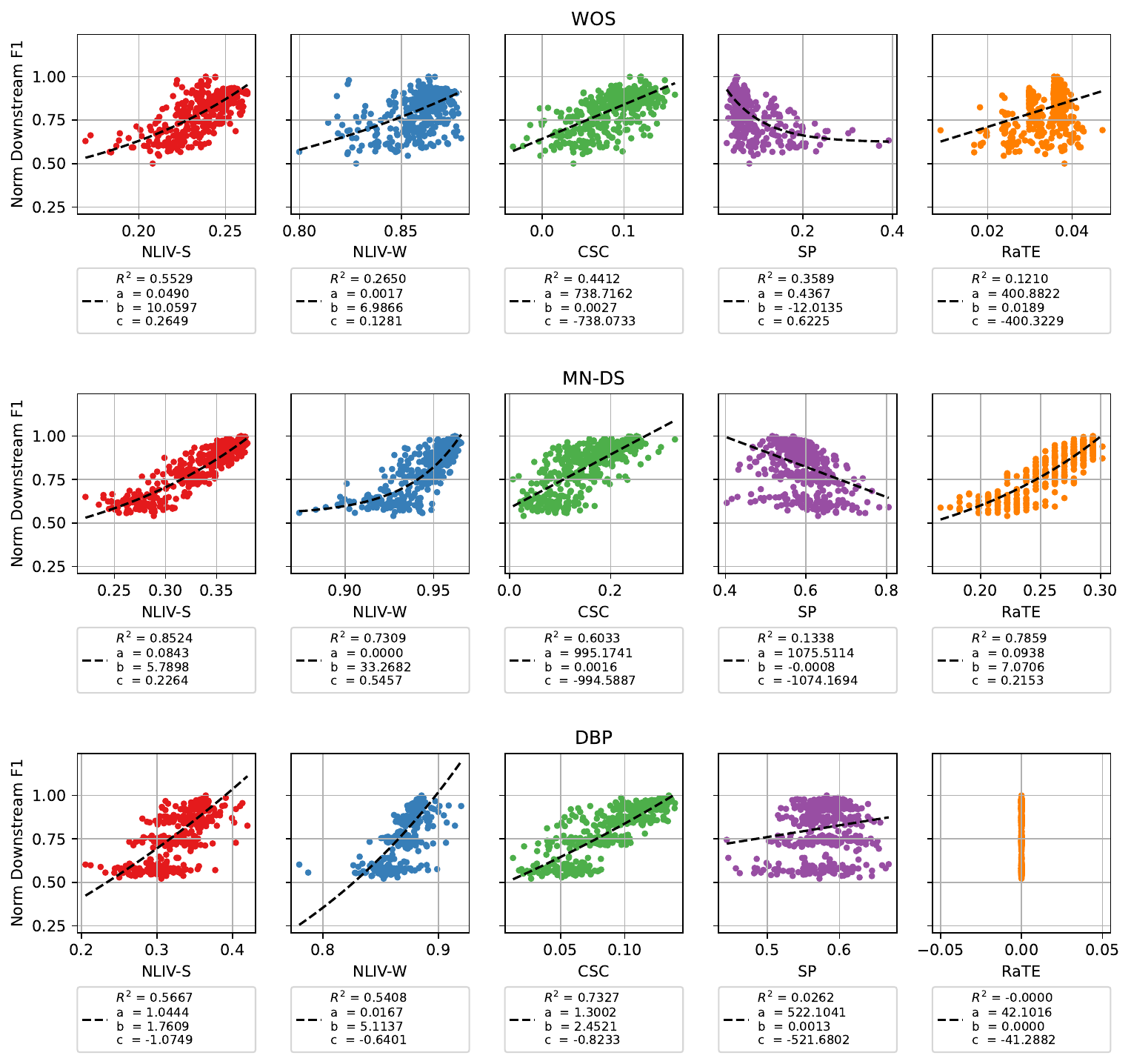}
\caption{Scatter plots of metric scores vs. F1 on the downstream task including fitted exponentials of form $y= a \cdot e^{bx} + c$.}
\label{fig:scatter_downstream_all}
\end{figure*}

For intuition, Fig.~\ref{fig:scatter_all_weighted} and Fig.~\ref{fig:scatter_all_unweighted} show scatterplots of the metrics for the collected samples of mutated taxonomies. Fig.~\ref{fig:scatter_downstream_all} shows the downstream scores vs. downstream F1, including the fitted exponential curves with their parameterizations.

\subsection{Approximating Parent-Child Likelihoods}
\label{app:approx_perf}
\begin{figure*}[!b]
\centering
\includegraphics[width=0.9\textwidth]{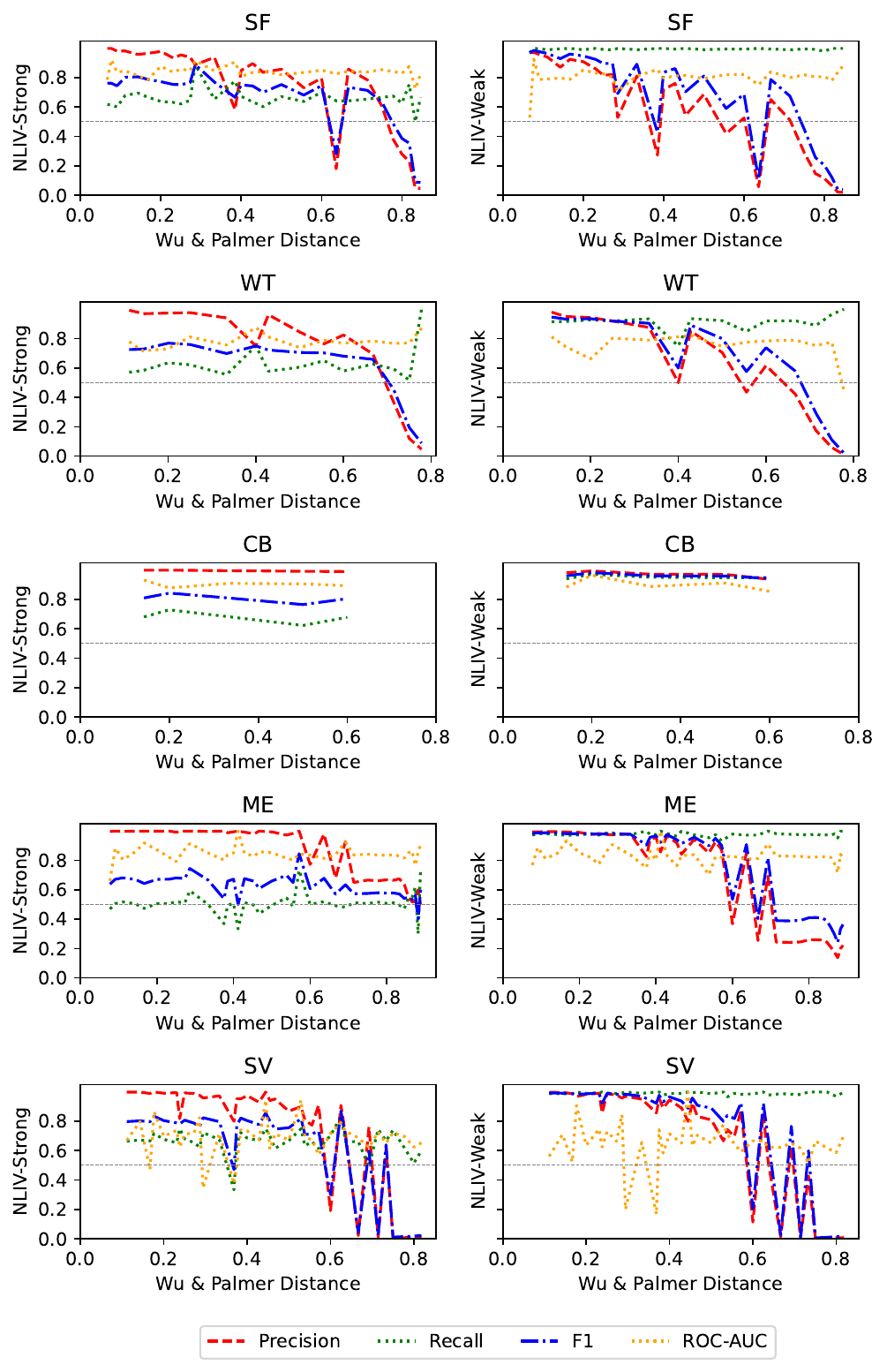}
\caption{Performance of NLI for estimating edge probabilities measured in binary F1, precision and recall and ROC-AUC by moving over windows along the \gls{WPS} of the concepts pairs of the edge. We randomly sample $|\E|$ negative pairs by randomly combining concepts without any common ancestor (except the root).}
\label{fig:nliv_perf_all}
\end{figure*}

Tab.~\ref{tab:approx_perf} shows the performance on the binary classification of parent-child edges for all datasets. Fig.~\ref{fig:nliv_perf_all} shows the performance given the taxonomic distance between the potential parent-child edges in the ground truth taxonomy. 

\paragraph{LLMs for NLI}
In preliminary experiments using a setting similar to Fig.~\ref{fig:nliv_perf_all},
prompting \texttt{Llama-3.1-8B-Instruct} \citep{grattafiori2024llama3herdmodels}
for NLI yielded comparable performance to NLI-trained models, and GPT-4o \citep{openai2024gpt4ocard} was only slightly better.
We did not consider these improvements worth the extra compute.
Obtaining continuous probabilities from LLMs would also require additional calibration,
which we avoided to keep the method simple and general.

\end{document}